\documentclass{article}
\usepackage{erlt_preprint,times}

\newcommand{\paperpdfauthor}{Dehao Huang, Jianbang Liu, Jianpan Gao, Chao Tang, Zilang Cen, Zedong Dan, Jiaheng Wang, Tingguang Li, Yue Wang, Hong Zhang}
\newcommand{\paperbibliographystyle}{erlt_natbib}
\newcommand{\paperprojectpage}{Project~page: \url{https://red0orange.github.io/eRLT_site/}.}
\newcommand{\paperleadershipfootnote}{%
  \begingroup
  \renewcommand{\thefootnote}{\fnsymbol{footnote}}%
  \footnotetext[2]{Project Leader.}%
  \endgroup}

\usepackage{amsmath,amsfonts,bm}

\def\eqref#1{equation~\ref{#1}}
\def\1{\bm{1}}

\DeclareMathAlphabet{\mathsfit}{\encodingdefault}{\sfdefault}{m}{sl}
\SetMathAlphabet{\mathsfit}{bold}{\encodingdefault}{\sfdefault}{bx}{n}

\usepackage{booktabs}
\usepackage{graphicx}
\usepackage{placeins}
\usepackage{float}
\usepackage{tabularx}
\usepackage{hyperref}
\usepackage{url}

\hypersetup{
  pdftitle={eRLT: Efficient VLA Reinforcement Learning via Action-Relevant Token Routing},
  pdfauthor={\paperpdfauthor}
}

\AtBeginDocument{
  \setlength{\abovedisplayskip}{3pt plus 1pt minus 1pt}
  \setlength{\belowdisplayskip}{3pt plus 1pt minus 1pt}
  \setlength{\abovedisplayshortskip}{2pt}
  \setlength{\belowdisplayshortskip}{4pt}
}

\title{eRLT: Efficient VLA Reinforcement Learning via Action-Relevant Token Routing}

\author{
\parbox{0.98\textwidth}{\centering
{\small
\mbox{Dehao Huang\textsuperscript{1,2,3}}\quad
\mbox{Jianbang Liu\textsuperscript{3}}\enspace
\mbox{Jianpan Gao\textsuperscript{3}}\enspace
\mbox{Chao Tang\textsuperscript{3}}\enspace
\mbox{Zilang Cen\textsuperscript{4,2}}\enspace
\mbox{Zedong Dan\textsuperscript{5,2}}\\[0.28em]
\mbox{Jiaheng Wang\textsuperscript{3}}\quad
\mbox{Tingguang Li\textsuperscript{3,\(\dagger\)}}\quad
\mbox{Yue Wang\textsuperscript{2}}\quad
\mbox{Hong Zhang\textsuperscript{1}}}\\[0.75em]
{\normalfont\small
\textsuperscript{1}\,Southern University of Science and Technology, Shenzhen, China.\\[0.08em]
\textsuperscript{2}\,Beijing Zhongguancun Academy, Beijing, China.\qquad
\textsuperscript{3}\,Samsung Robotics eXperience.\\[0.08em]
\textsuperscript{4}\,Wuhan University, Wuhan, China.\qquad
\textsuperscript{5}\,Sun Yat-sen University, Guangzhou, China.}}
}
\begin{document}

\maketitle
\paperleadershipfootnote

\begin{abstract}
Vision-Language-Action (VLA) models provide strong behavioral priors for robotic manipulation, yet efficiently adapting them to downstream tasks remains challenging. Recent work addresses this challenge by adapting frozen VLAs through online reinforcement learning (RL), whose sample efficiency depends on the quality of the state representation used by the actor and critic. Existing methods construct such representations either with VLA-independent visual encoders or through fixed compression of internal VLA representations. Neither design explicitly extracts the task-specific action-relevant VLA features most useful for downstream action refinement and action-value estimation, therefore limiting sample efficiency. To address this limitation, we introduce eRLT, which constructs an effective state representation by routing task-specific action-relevant information across both tokens and layers of the frozen VLA. Specifically, learned routing tokens dynamically aggregate visual-language features at multiple depths, while a lightweight layer router combines these summaries into a fixed-dimensional RL token. The routing module is initialized using expert demonstrations to capture features predictive of expert actions and then refined using critic feedback from online interactions for action-value estimation. Across seven LIBERO and RoboTwin tasks, eRLT improves mean normalized learning-curve AUC by up to $23.7\%$ over representative baselines. Real-robot experiments on USB connector insertion and motherboard ribbon-cable insertion further show AUC improvements of $108.9\%$ and $46.7\%$, respectively, over the strongest baseline. \paperprojectpage
\end{abstract}

\section{Introduction}
Vision-language-action (VLA) models inherit broad visual and semantic knowledge from internet-scale pretraining and learn generalist control from diverse robot data, enabling generalization across tasks and environments \citep{kim2025openvla,black2026pi0}. However, achieving high reliability on a specific downstream task often requires further adaptation. Supervised fine-tuning (SFT) is effective, but heavily relies on collecting additional high-quality task-specific demonstrations, whose coverage of states encountered during deployment is inherently limited. Online reinforcement learning (RL) provides a complementary approach by collecting experience from the policy's own deployment distribution and learning from both successful and failed interactions \citep{physicalintelligence2025pi06,xu2026rltoken,wagenmaker2025dsrl,xiao2025pld, zhu2026learning, gao2026smoothrl}. In this work, we focus on lightweight online RL methods that keep the pretrained VLA frozen and adapt its behavior through a small actor-critic.

The success of lightweight online RL depends critically on the state representation supplied to the actor and critic. A VLA trained for generalist control encodes rich information about visual observations, language instructions, robot states, and actions. For task-specific adaptation, an effective state representation should retain the task-specific action-relevant information needed for action refinement and action-value estimation, including robot-object relations that determine appropriate corrections and task-progress cues that determine the expected return of those corrections. Such information is distributed across the VLA's tokens and layers, and the most useful features vary across tasks. Existing methods construct the state representation either with visual encoders independent of the VLA \citep{wagenmaker2025dsrl,xiao2025pld} or by compressing representations from a predetermined layer and token set \citep{kowshik2026midas,xu2026rltoken}. The former constructs the state representation outside the VLA and therefore cannot directly reuse its internal task-specific action-relevant features. The latter reuses VLA features, but restricts feature extraction to a predetermined layer and token set regardless of which features are most useful for the target task. Neither design explicitly extracts the task-specific action-relevant VLA features most useful for downstream action refinement and action-value estimation, limiting sample efficiency under constrained online interaction budgets.

\begin{figure*}[t]
\centering
\includegraphics[width=0.85\textwidth, trim={0.5mm 3mm 0.5mm 3mm}, clip]{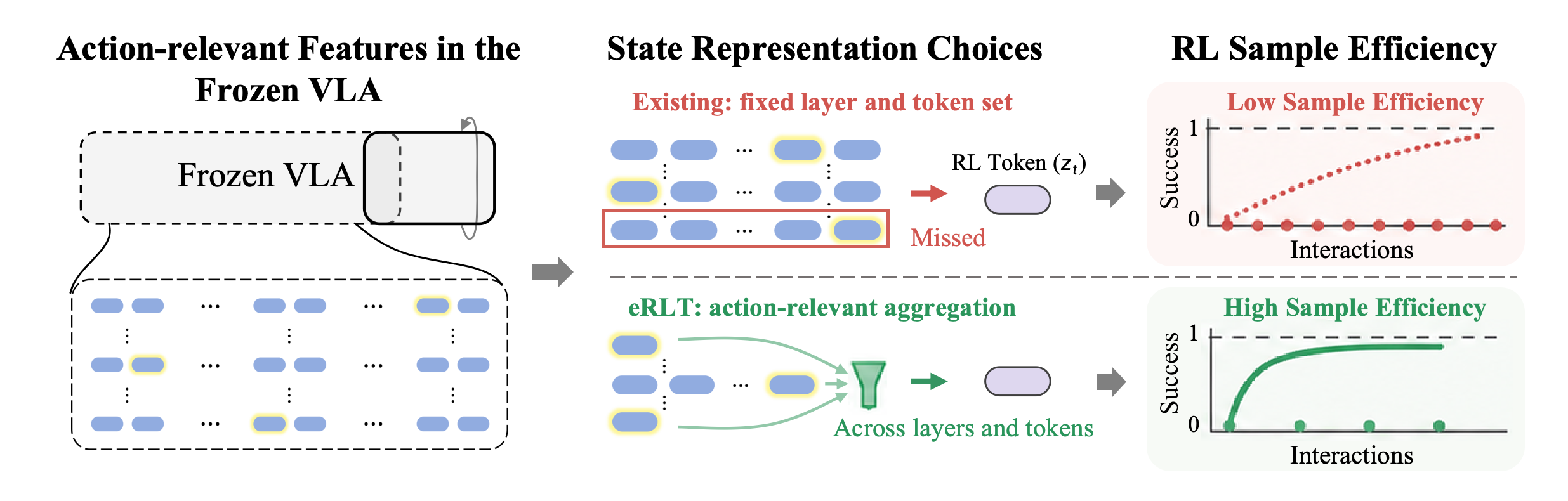}
\caption{\textbf{Motivation for task-specific action-relevant RL token construction.} The task-specific action-relevant information needed for action refinement and action-value estimation, which is represented by yellow highlight, is distributed across the tokens and layers of a frozen VLA. Existing methods that use fixed layers and tokens can miss features useful for the downstream task, whereas eRLT aggregates them across both dimensions to improve online RL sample efficiency.}
\label{fig:motivation}
\vspace{-1pt}
\end{figure*}

To address this limitation, we introduce eRLT, which constructs an effective state representation by routing task-specific action-relevant information across the tokens and layers of the frozen VLA. At each selected VLM layer, learned routing tokens dynamically aggregate visual-language features, while a lightweight layer router combines these layer-wise summaries into a fixed-dimensional RL token. The routing module is firstly initialized by predicting expert action chunks from existing VLA SFT demonstrations, encouraging the RL token to retain information useful for action refinement. Then, during online RL stage, the critic's Bellman objective further refines the routing module using online transitions, adapting the retained information for action-value estimation. The VLA remains frozen throughout both stages. By focusing the RL token on task-specific action-relevant VLA features, eRLT facilitates action refinement and action-value estimation, thereby supporting sample-efficient actor-critic learning from limited online interactions.

We evaluate eRLT on manipulation tasks from LIBERO \citep{liu2023libero} and RoboTwin \citep{chen2025robotwin}, as well as two real-world high-precision manipulation tasks: USB connector insertion and motherboard ribbon-cable insertion. Across the seven simulation tasks, eRLT improves mean normalized learning-curve AUC by $23.7\%$ relative to baselines with fixed VLA feature compression and by $7.1\%$ relative to baselines with VLA-independent state encoders. On the two real-world tasks, eRLT improves learning-curve AUC by $108.9\%$ and $46.7\%$, respectively, relative to the strongest baseline, and on USB connector insertion raises the final-window success rate from $50\%$ to $90\%$ under the same interaction budget. Ablations further validate the complementary roles of layer-wise routing, token-wise routing, task-specific action-relevant initialization, and online adaptation.

In summary, our contributions are three-fold:
\begin{itemize}
    \setlength{\topsep}{2pt}
    \setlength{\itemsep}{0pt}
    \setlength{\parsep}{0pt}
    \setlength{\parskip}{0pt}
    \setlength{\partopsep}{0pt}
    \item We provide controlled evidence that the VLA layers and tokens most useful for online RL vary across downstream tasks.
    \item We propose eRLT, which constructs an effective RL token by routing task-specific action-relevant information across VLA tokens and layers. The routing mechanism is initialized via expert action prediction and subsequently refined online using critic feedback.
    \item We evaluate eRLT across LIBERO, RoboTwin, and two real-world precision high-precision manipulation tasks, demonstrating improved sample efficiency over existing state representations.
\end{itemize}

\section{Related Work}

\subsection{Lightweight Online Reinforcement Learning for VLA Models}
Lightweight online RL adapts a pretrained policy through small trainable modules while preserving its behavioral prior. Early residual RL learns corrections for conventional controllers \mbox{\citep{johannink2018residualrl,silver2018residualpolicy}}; recent VLA methods instead keep the pretrained policy frozen and train a lightweight actor to correct its actions or steer its generative process \citep{xiao2025pld,wagenmaker2025dsrl,li2025grrl,xu2026rltoken,zhu2026learning,gao2026smoothrl,kowshik2026midas,zheng2026torlvla}. Some further incorporate human guidance or improved initialization \citep{lu2026unisteerunifiednoisesteering,frs}. eRLT adopts this frozen-VLA actor-critic framework and studies how to construct the state representation supplied to the actor and critic from the VLA's internal representations.

Within this framework, the actor and critic must infer which decisions lead to higher returns from an effective state representation, making the task-specific action-relevant information retained in this representation central to sample-efficient learning. Existing methods construct the state representation either with VLA-independent visual encoders or by compressing a predetermined layer and token set. DSRL and PLD use the former design \citep{wagenmaker2025dsrl,xiao2025pld}; MiDAS mean-pools internal VLM tokens, while RLT compresses final-layer tokens using reconstruction \citep{kowshik2026midas,xu2026rltoken}. The former constructs state representations separately from those already available in the VLA, while the latter uses only prescribed layers and tokens. Both designs therefore lack a mechanism for selecting the internal VLA information most relevant to task-specific action refinement and action-value estimation, limiting sample efficiency under constrained online interaction budgets. eRLT instead learns to aggregate task-specific action-relevant information across VLA tokens and layers into an effective RL token.

\subsection{Internal VLA Representations and Feature Aggregation}

Frozen VLA representations encode object relations, robot and action states, task progress, failures, transitions, and future success \citep{lu2025probing,gu2025safe,bhardwaj2026taskprogress,zhang2026frozenvlasuccess,molinari2025emergent}. Intervention studies further show that motion- and action-related features can influence robot behavior \citep{haon2025mechanistic,buurmeijer2026observing,swann2026sparsevla,grant2026features}. These signals are distributed: different layers provide complementary representations \citep{wang2026vlaadapter,yao2024denseconnector,cao2024mmfuser,lin2025multilayer,shi2026vlatrace}, while selective token aggregation can improve action generation \citep{xiao2026avavla,ji2026actionqformer}. These findings support eRLT's use of frozen internal VLA representations as the source of the RL token supplied to the actor and critic.

Feature aggregation methods extract such distributed information either by compressing tokens within a layer or by combining representations across layers. Learned tokens and queries compress visual or multimodal features within a layer \citep{ryoo2021tokenlearner,li2023blip2,li2025tokenpacker}, and recent VLA methods use related mechanisms for action generation \citep{xiao2026avavla,ji2026actionqformer,kim2026progvla}. Other methods combine hidden representations across depths \citep{yao2024denseconnector,cao2024mmfuser,lin2025multilayer,wang2026vlaadapter}. These representations target visual-language understanding or VLA action generation; eRLT adapts both forms of aggregation to construct a fixed-dimensional RL token that retains task-specific action-relevant information for the actor and critic.

\section{Preliminaries and Problem Formulation}

\subsection{Lightweight Online RL with a Frozen VLA}

At each policy step $t$, the robot receives an observation $o_t=(I_t,x_t,c)$, where $I_t$ denotes one or more camera views, $x_t$ is the robot proprioceptive state, and $c$ is the language instruction. A pretrained VLA provides a strong behavioral prior for generating task-relevant robot actions. Lightweight online RL keeps the VLA frozen as the action prior and generator and improves its behavior by training only lightweight modules around it, such as an actor and critic. The frozen VLA and the trainable actor jointly produce an executable action chunk from the current observation and the actor's RL state. Different lightweight RL algorithms may combine the actor and VLA in different ways; their concrete implementations are specified in the experimental setup.

Online training alternates between environment interaction and actor-critic updates. After executing an action chunk, the robot receives a reward $R_t$, the next observation $o_{t+1}$, and a termination indicator $d_t$. The interaction is stored in a replay buffer and repeatedly reused for off-policy learning \citep{haarnoja2019sac}. The critic learns which decisions of the lightweight actor lead to high long-term return, and the actor is updated to favor decisions assigned higher value by the critic. Thus, the lightweight modules improve from both successful and failed trials, while the pretrained VLA remains unchanged throughout online learning. The generic interaction and learning loop is shown in Appendix~\ref{app:lightweight-online-rl}.

\subsection{RL State Representation}

The formulation above leaves the construction of the lightweight controller to the chosen online RL algorithm. The component shared across these algorithms, and the focus of this work, is the state representation supplied to the actor and critic. Our goal is to construct an effective state representation from the frozen VLA's multi-layer token representations while preserving the task-specific action-relevant information needed for action refinement and long-horizon action-value estimation. We denote the resulting representation by $z_t$, which summarizes information distributed across tokens and layers. The actor and critic combine $z_t$ with proprioception through separate adapters to form their RL states. Within each experimental setting, we keep the frozen VLA, online RL algorithm, and action-generation procedure fixed, and vary only how the RL token is constructed and trained. Section~\ref{sec:method} introduces the proposed task-specific action-relevant multi-layer token routing method.

\section{Method}
\label{sec:method}

\subsection{Motivating Study: Layer and Token Choices}
\label{sec:motivating-study}

We first examine how the VLA layers and tokens used to construct the RL token affect online RL. We use RL Token (RLT) \citep{xu2026rltoken} as a controlled baseline on two LIBERO tasks \citep{liu2023libero}, keeping the frozen VLA, actor-critic architecture, online RL procedure, and interaction budget fixed. We vary only the VLA features supplied to RLT: the source layer with all tokens retained, or the token subset with the final layer retained. We report mean normalized success-rate AUC over three seeds; additional details are provided in Appendix~\ref{app:motivating-study-details}.

\begin{figure*}[t]
\centering
\includegraphics[width=0.95\textwidth, trim={1mm 2mm 1mm 5mm}, clip]{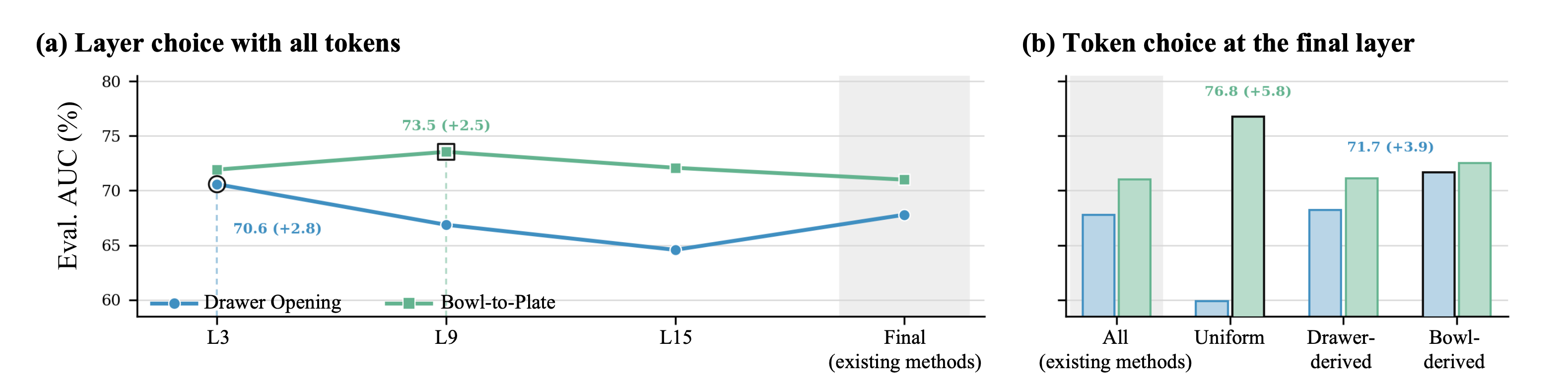}
\caption{\textbf{Layer and token choices affect online RL.} Controlled RLT comparisons on Drawer Opening and Bowl-to-Plate. (a) With all tokens retained, the best source layer differs across tasks. (b) At the final layer, selected token subsets can outperform all tokens, while the best subset also varies across tasks. Values report mean normalized success-rate AUC over three seeds; annotations show absolute improvements over the Final/All baselines.}
\label{fig:rlt-feature-selection}
\end{figure*}

Figure~\ref{fig:rlt-feature-selection} shows that online RL is sensitive to both choices. The best source layer differs across tasks, improving over the final layer by $2.8$ and $2.5$ percentage points on Drawer Opening and Bowl-to-Plate, respectively. Selected token subsets also outperform all tokens by $3.9$ and $5.8$ points, with a different subset performing best on each task. Thus, the VLA features most useful for online RL vary across tokens, layers, and downstream tasks, motivating eRLT to learn how they are aggregated.

\subsection{Overview}

Motivated by this study, eRLT uses learned routing tokens to aggregate multimodal VLA features at $m$ selected VLM layers $\ell_1,\ldots,\ell_m$ and a lightweight layer router to combine them into the RL token $z_t$. Token summaries depend on the current input, whereas layer weights are shared within each adaptation task. The routing parameters $\eta$ comprise the routing-token embeddings $\eta_{\mathrm{tok}}$ and layer logits $\eta_{\mathrm{lay}}$.

\begin{figure*}[t]
\centering
\includegraphics[width=0.96\textwidth, trim={1.5mm 3mm 1mm 2mm}, clip]{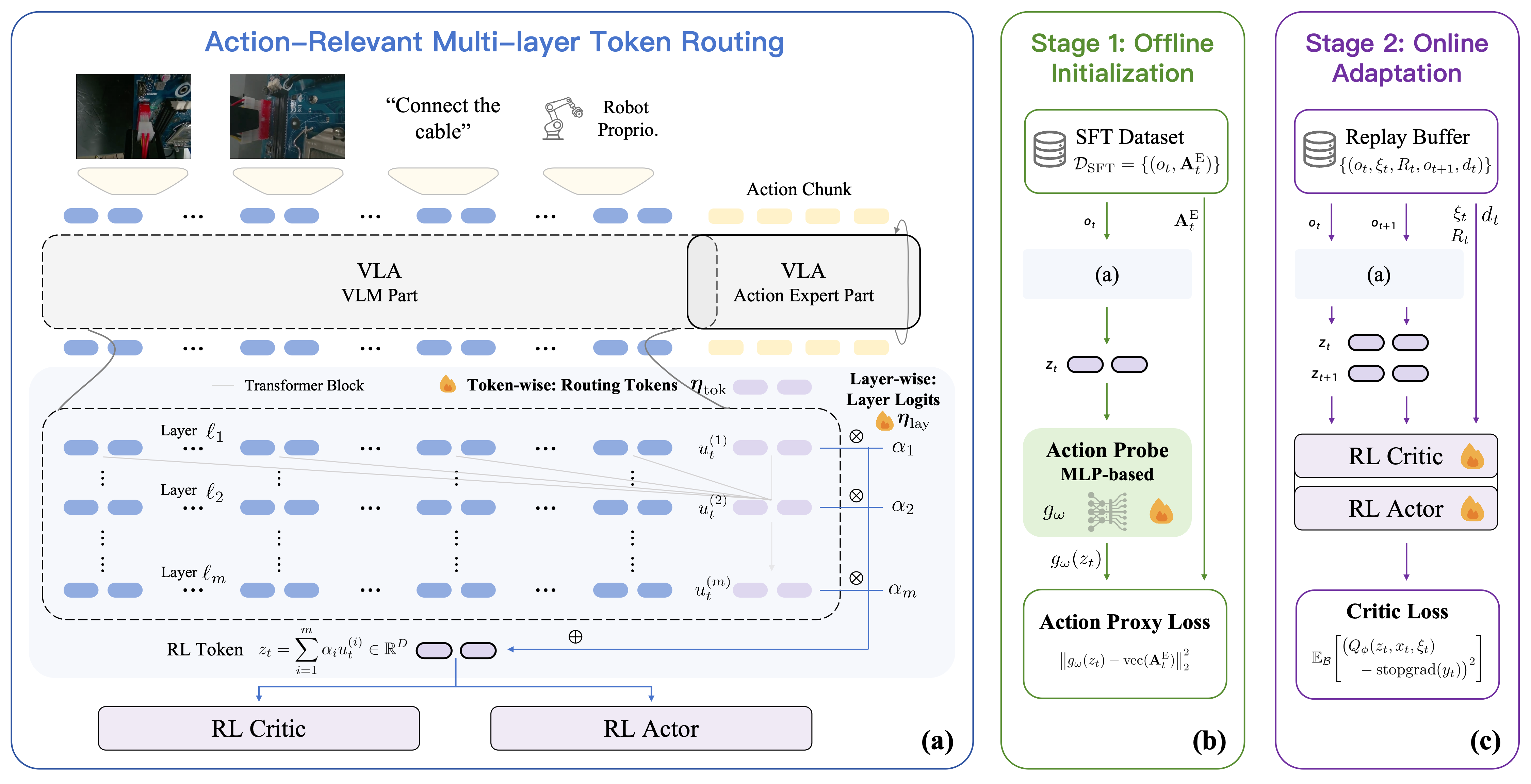}
\caption{\textbf{Overview of eRLT.} (a) Task-specific action-relevant multi-layer token routing constructs an RL token using learnable routing tokens and layer-wise routing. (b) Expert action prediction teaches the RL token to preserve differences needed for action refinement. (c) The critic uses successful and failed transitions to emphasize cues needed for action-value estimation.}
\label{fig:method-overview}
\end{figure*}

What the RL token retains is determined by two sources of supervision. Before online interaction, expert action prediction teaches it to preserve differences needed for action refinement. During online RL, successful and failed transitions reveal which differences affect task progress and outcomes; the critic objective adapts the routing parameters accordingly for action-value estimation. The first stage provides an action-predictive starting point from existing demonstrations, while the second adapts it to situations encountered by the current policy. The VLA remains frozen throughout both stages. Figure~\ref{fig:method-overview} summarizes the routing module and both training stages.

\vspace{-8pt}

\subsection{Task-Specific Action-Relevant Multi-Layer Token Routing}
\label{sec:multi-layer-token-routing}

\paragraph{Token-wise routing.}
As shown in Figure~\ref{fig:method-overview}(a), let $E_t$ denote the multimodal prefix embeddings produced from the current images, proprioception, and instruction. Task-specific cues can appear at different positions as the scene and instruction change. We append $K$ routing tokens $\eta_{\mathrm{tok}}$ to $E_t$ in a separate forward pass. Their attention changes with the current input, allowing their hidden states to gather different prefix information at each policy step. The action-prediction and critic objectives optimize these embeddings for action refinement and action-value estimation. This forward pass is separate from VLA action generation, so the routing tokens do not alter the action proposals produced by the frozen VLA.

For each selected layer $\ell_i$, we take the hidden states at the $K$ routing-token positions and mean pool them:
\begin{equation}
u_t^{(i)}=
\frac{1}{K}\sum_{k=1}^{K}
\left[
F_{\mathrm{VLM}}^{(\ell_i)}
([E_t;\eta_{\mathrm{tok}}])
\right]_{\mathrm{route},k}.
\label{eq:token-routing-readout}
\end{equation}
Here, $F_{\mathrm{VLM}}^{(\ell_i)}$ maps the input embedding sequence to the sequence of token representations at layer $\ell_i$ of the frozen VLM, $[\cdot;\cdot]$ denotes concatenation along the token dimension, and $[\cdot]_{\mathrm{route},k}$ extracts the hidden state at the $k$-th appended routing-token position. Each selected layer therefore contributes one summary $\smash[t]{u_t^{(i)}}$ regardless of the number of prefix or routing tokens.

\paragraph{Layer-wise routing.}
The same routing-token positions are read at every selected depth, so each $u_t^{(i)}$ summarizes their hidden states at a different stage of VLA processing. Together, these representations summarize the same input at several depths. To adapt their relative contributions to each task, we learn layer weights shared across observations within each adaptation task, normalized as $\alpha=\operatorname{softmax}(\eta_{\mathrm{lay}}/\tau)$, where $\tau$ controls their concentration. The RL token is
\begin{equation}
z_t=\sum_{i=1}^{m}\alpha_i u_t^{(i)},
\label{eq:layer-routing-readout}
\end{equation}
Token-wise routing determines what information is gathered for the current input, whereas layer-wise routing determines which VLA depths contribute to the RL token $z_t$.

\subsection{Task-Specific Action-Relevant Initialization}
\label{sec:action-relevant-initialization}

Before online interaction, eRLT reuses the VLA SFT dataset $\mathcal{D}_{\mathrm{SFT}}$ to teach both routing stages which internal differences matter for expert actions, without additional environment experience. As shown in Figure~\ref{fig:method-overview}(b), a lightweight action probe $g_\omega$ predicts the expert action chunk $\mathbf{A}_t^{\mathrm{E}}$ from the RL token using
\begin{equation}
\mathcal{L}_{\mathrm{proxy}}
=
\mathbb{E}_{(o_t,\mathbf{A}_t^{\mathrm{E}})\sim\mathcal{D}_{\mathrm{SFT}}}
\left[
\left\|
g_\omega(z_t)-\operatorname{vec}(\mathbf{A}_t^{\mathrm{E}})
\right\|_2^2
\right].
\label{eq:action-proxy-loss}
\end{equation}
When two observations require different expert action chunks, the probe can predict both only if $z_t$ preserves the difference between them. Because the probe receives only $z_t$, minimizing Equation~\ref{eq:action-proxy-loss} teaches both routing stages which information is needed for action refinement. This differs from reconstruction, which asks the RL token to preserve the full VLA token content. The VLA remains frozen while $\eta$ and $\omega$ are updated. We select $\eta$ by validation action-prediction error and remove the probe before online RL.

\subsection{Critic-Driven Online Adaptation}
\label{sec:critic-driven-adaptation}

The initialization above captures information predictive of expert actions. Online transitions provide a complementary criterion for action-value estimation: when similar interactions lead to different progress or outcomes, reducing critic error requires $z_t$ to expose cues that explain this difference. As shown in Figure~\ref{fig:method-overview}(c), we therefore backpropagate the critic objective into the routing parameters. For example, in USB connector insertion, the critic must distinguish small positional and angular misalignments because they determine whether insertion succeeds.

Let $\xi_t$ denote the action selected by the lightweight actor. For each transition $(o_t,\xi_t,R_t,o_{t+1},d_t)$ sampled from the replay buffer $\mathcal{B}$, we compute the TD target $y_t$ using the target routing module, actor, and critic. We optimize the critic and routing parameters with
\begin{equation}
\mathcal{L}_{Q}
=
\mathbb{E}_{\mathcal{B}}
\left[
\left(
Q_\phi(z_t,x_t,\xi_t)
-\operatorname{stopgrad}(y_t)
\right)^2
\right].
\label{eq:critic-routing-loss}
\end{equation}
We backpropagate the critic loss through $z_t$ to update the routing parameters $\eta$ together with the critic parameters $\phi$. Actor updates use the current RL token but detach it, so they do not change how the RL token is constructed. Target copies of the actor, critic, and routing module follow their online counterparts through Polyak averaging, allowing the RL tokens used in the Bellman target to change gradually. To limit computation and the rate of representation change, routing parameters are updated periodically. The replay buffer stores the original observations, from which we recompute the current and next RL tokens during routing updates. Appendix~\ref{app:routing-optimization} provides the complete gradient flow and implementation details.

\section{Experiments}
\label{sec:experiments}

We design our experiments to answer three questions: (1) Does eRLT improve the sample efficiency of lightweight online VLA adaptation across different manipulation settings? (2) How do its routing components and two-stage training strategy contribute to the improvement? (3) Does the resulting RL token support efficient adaptation on real-world precision manipulation tasks?

\subsection{Experimental Setup}

\noindent\textbf{Simulation benchmarks.}
We evaluate eRLT on LIBERO \citep{liu2023libero} and RoboTwin \citep{chen2025robotwin}, which feature diverse object layouts, language instructions, and manipulation scenarios. The four LIBERO tasks are Drawer Opening, Bowl-to-Plate, Cabinet-Bowl-to-Plate, and Two-Moka-Pots, and the three RoboTwin tasks are Adjust Bottle, Move Can--Pot, and Place Container Plate. Together, they cover spatial reasoning, object relocation, and multi-stage manipulation.

\noindent\textbf{Real-world tasks.}
We further evaluate eRLT on two precision manipulation tasks: USB connector insertion and motherboard ribbon-cable insertion. Experiments use the left arm of an RB-Y1 robot and two wrist-mounted cameras. Both tasks require precise visual alignment followed by contact-rich insertion. Starting from the official $\pi_{0.5}$ base checkpoint, the task-specific VLAs are fine-tuned on 99 USB and 800 ribbon-cable teleoperated demonstrations, respectively. To vary the initial end-effector pose, each reset independently perturbs the seven left-arm joint targets with zero-mean Gaussian noise of standard deviation $0.01$\,rad per joint. Online RL refines the VLA reference behavior in adapted episodes. Every method uses 80 trajectories on USB insertion and 90 trajectories on ribbon-cable insertion under the same human-in-the-loop protocol.

\noindent\textbf{Compared methods.}
We compare three state representations while keeping the frozen VLA and online RL procedure fixed within each setting. \textbf{DSRL$^\ast$} uses a VLA-independent visual encoder \citep{wagenmaker2025dsrl}. \textbf{RLT$^\ast$} compresses prescribed final-layer VLA tokens into an RL token through reconstruction and keeps the RL token fixed during online RL \citep{xu2026rltoken}. \textbf{eRLT} instead constructs the proposed task-specific action-relevant RL token through multi-layer token routing. The superscript $\ast$ denotes use of only the corresponding state representation in our shared online RL framework. We also report the \textbf{Frozen VLA} before adaptation.

\noindent\textbf{Evaluation protocol and metrics.}
Our primary measure of online sample efficiency is the normalized area under the success-rate learning curve (AUC), computed by trapezoidal integration and normalized by the evaluation horizon. It reflects average success over a fixed interaction budget, rewarding methods that reach high success rates earlier and sustain them. We omit final success rates in simulation because the methods eventually reach similar levels. For the real-world experiments, we report both normalized AUC and success rate in the final 10-episode window; the Frozen VLA reference is evaluated over 10 trials.

\noindent\textbf{Implementation details.}
Simulation uses a DSRL-style off-policy actor-critic that steers the frozen VLA in its latent action space \citep{wagenmaker2025dsrl}, whereas the real-world study uses the RLT pipeline \citep{xu2026rltoken}. Within each setting, all variants share the VLA checkpoint, actor and critic architectures, replay data, reward, update schedule, and interaction budget; only the state representation varies. RLT and eRLT reuse the task-specific SFT demonstrations for offline initialization, and eRLT uses one routing token unless otherwise stated. Appendix~\ref{app:experimental-setup} provides task visualizations and complete implementation details, including the real-world hardware and collection protocol.

\subsection{Main Simulation Results}

\begin{figure*}[t]
\centering
\includegraphics[width=0.92\textwidth, trim={1mm 5mm 1mm 5mm}, clip]{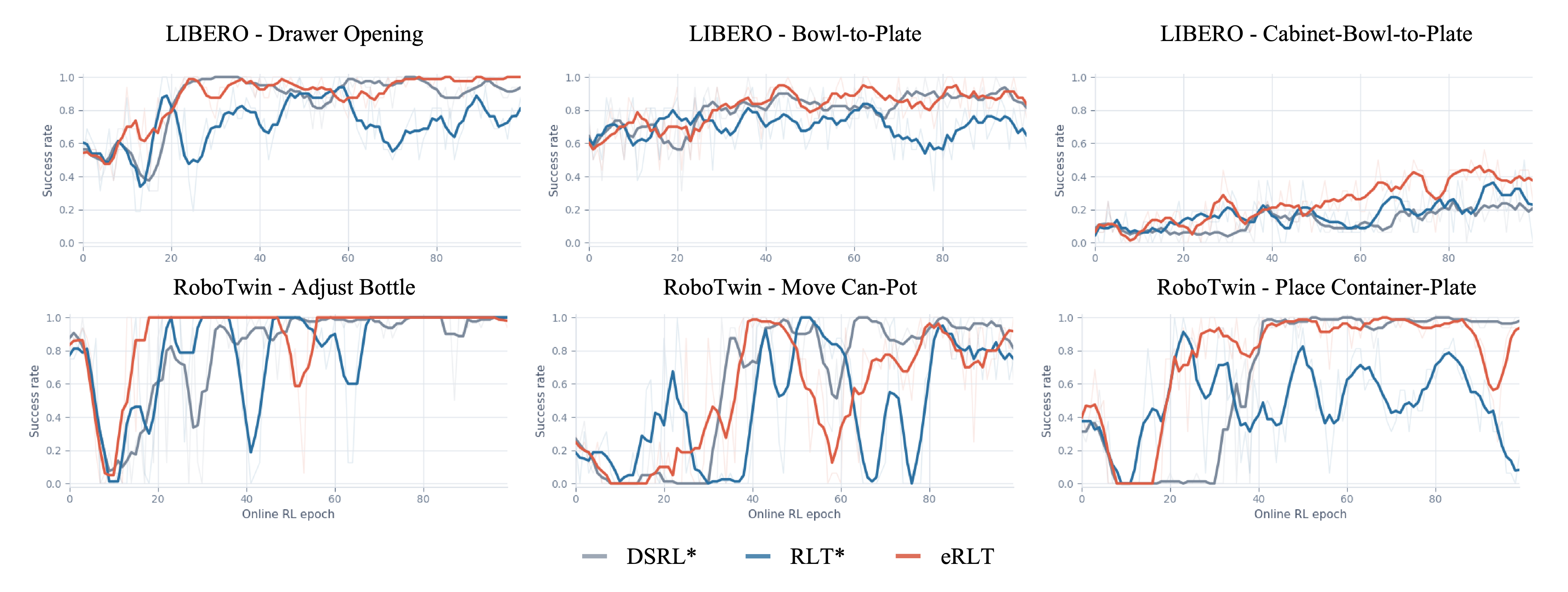}
\caption{\textbf{Online learning curves in simulation.} Evaluation success rate over online RL epochs on three representative LIBERO tasks (top) and three RoboTwin tasks (bottom). Dark curves show the smoothed trends, and light curves show the corresponding raw evaluations.}
\label{fig:main-learning-curves}
\end{figure*}

Table~\ref{tab:main-simulation} summarizes normalized learning-curve AUC on all seven simulation tasks, while Figure~\ref{fig:main-learning-curves} shows the corresponding learning dynamics on six representative tasks. Before online adaptation, the frozen VLA achieves success rates of $74\%$, $78\%$, $12\%$, and $14\%$ on the four LIBERO tasks and $60\%$, $12\%$, and $72\%$ on the three RoboTwin tasks, respectively, over 50 trials per task. After online RL adaptation, eRLT substantially improves upon these frozen-VLA success rates on all seven tasks. Overall, eRLT achieves the strongest aggregate performance across both benchmarks, outperforming RLT$^\ast$ on every task and DSRL$^\ast$ on six of seven. Because all variants share the same VLA and online RL procedure, these gains isolate the effect of the state representation and show that a task-specific action-relevant RL token constructed from internal VLA features improves sample efficiency across environments and robot embodiments.

\begin{table}[b]
\caption{\textbf{Main simulation results.} Mean normalized success-rate AUC on four LIBERO and three RoboTwin tasks. The final column averages all seven tasks. Best and second-best results are bold and underlined, respectively.}
\label{tab:main-simulation}
\vspace{2pt}
\centering
\scriptsize
\renewcommand{\arraystretch}{1.08}
\begin{tabular*}{\columnwidth}{@{\extracolsep{\fill}}lcccccccc@{}}
\toprule
Method & \multicolumn{4}{c}{LIBERO} & \multicolumn{3}{c}{RoboTwin} & \multicolumn{1}{c}{All tasks} \\
\cmidrule(lr){2-5}\cmidrule(lr){6-8}\cmidrule(l){9-9}
& \shortstack{Drawer\\Opening} & \shortstack{Bowl-to-\\Plate} & \shortstack{Cabinet-Bowl\\to-Plate} & \shortstack{Two-Moka\\Pots} & \shortstack{Adjust\\Bottle} & \shortstack{Move Can--\\Pot} & \shortstack{Place Container\\Plate} & Mean \\
\midrule
DSRL$^\ast$ & $\underline{0.856}$ & $\underline{0.801}$ & $0.130$ & $\underline{0.244}$ & $\underline{0.804}$ & $\mathbf{0.608}$ & $\underline{0.649}$ & $\underline{0.585}$ \\
RLT$^\ast$ & $0.709$ & $0.707$ & $\underline{0.171}$ & $0.207$ & $0.797$ & $0.461$ & $0.490$ & $0.506$ \\
\addlinespace[1pt]
\textbf{eRLT (ours)} & $\mathbf{0.881}$ & $\mathbf{0.818}$ & $\mathbf{0.247}$ & $\mathbf{0.264}$ & $\mathbf{0.897}$ & $\underline{0.523}$ & $\mathbf{0.752}$ & $\mathbf{0.626}$ \\
\bottomrule
\end{tabular*}
\vspace{-2pt}
\end{table}

\vspace{-5pt}

On LIBERO, eRLT achieves the best result on all four tasks: it reaches high success rates earlier on Drawer Opening and Bowl-to-Plate and makes greater progress on the more challenging Cabinet-Bowl-to-Plate task. On RoboTwin, the gains on Adjust Bottle and Place Container Plate demonstrate that the learned RL token remains effective across different object interactions and multi-stage manipulation. DSRL$^\ast$ performs better on Move Can--Pot, although eRLT remains stronger than RLT$^\ast$, indicating that the relative benefit of a state representation can still be task dependent. Since the methods eventually approach similar success rates, eRLT's consistently larger AUCs show that the state representation determines how quickly the actor-critic can exploit the frozen VLA prior. Its consistent improvement over RLT$^\ast$ further indicates that fixed compression of a predetermined layer and token set does not fully exploit the VLA's internal representations; the RL token must preserve task-specific action-relevant information for downstream action refinement and action-value estimation.

\subsection{Ablation Studies}

We conduct ablations on the Bowl-to-Plate task and report normalized learning-curve AUC. Table~\ref{tab:component-ablation} removes each routing or training component under the same VLA and actor-critic setup: token-wise routing is replaced by uniform token pooling, layer-wise routing by the final VLA layer, offline initialization by a randomly initialized routing module, and online adaptation by a frozen initialized routing module. Table~\ref{tab:routing-token-capacity} varies the number of routing tokens while keeping the output dimension fixed.

\begin{table}[h]
\centering
\begin{minipage}[t]{0.45\linewidth}
\centering
\caption{\textbf{Component ablations.} Each row removes one component of the task-specific action-relevant RL token.}
\label{tab:component-ablation}
\vspace{2pt}
\footnotesize
\setlength{\tabcolsep}{4pt}
\begin{tabular}{lc}
\toprule
Variant & AUC $\uparrow$ \\
\midrule
eRLT & 0.818 \\
w/o offline initialization & 0.762 \\
w/o online adaptation & 0.803 \\
w/o token-wise routing & 0.742 \\
w/o layer-wise routing & 0.797 \\
\bottomrule
\end{tabular}
\end{minipage}
\hfill
\begin{minipage}[t]{0.45\linewidth}
\centering
\caption{\textbf{Routing-token capacity.} The output dimension is fixed while the routing-token count varies.}
\label{tab:routing-token-capacity}
\vspace{2pt}
\footnotesize
\setlength{\tabcolsep}{6pt}
\begin{tabular}{cc}
\toprule
\shortstack{Number of routing\\tokens ($K$)} & AUC $\uparrow$ \\
\midrule
1 & 0.818 \\
2 & 0.823 \\
4 & 0.749 \\
8 & 0.793 \\
\bottomrule
\end{tabular}
\end{minipage}
\end{table}

Removing any component reduces AUC, with the largest degradation observed without token-wise routing. This suggests that adaptively extracting task-specific action-relevant cues is more effective than uniformly pooling the visual-language prefix. Offline initialization is the next most influential component, indicating that an action-predictive starting point reduces the burden on the critic, while layer-wise routing and online adaptation provide complementary refinement. Increasing the number of routing tokens does not improve performance monotonically. Additional routing tokens introduce more routing parameters and require coordination from limited feedback before mean pooling; this added learning burden can outweigh their capacity benefit, supporting a single routing token as the default.

\subsection{Real-World Experiments}

\begin{figure*}[t]
\centering
\includegraphics[width=0.92\textwidth, trim={0.5mm 5mm 0.5mm 5mm}, clip]{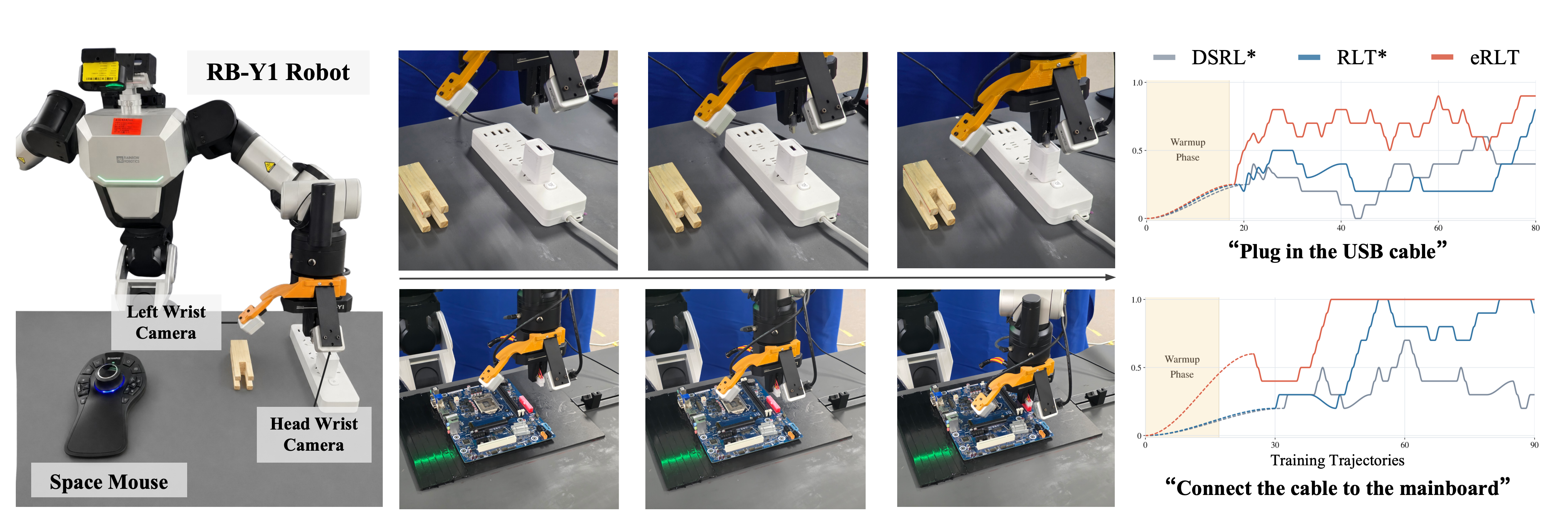}
\caption{\textbf{Real-world experimental setup and online learning curves.} The RB-Y1 robot setup (left), representative execution sequences for USB connector insertion (top center) and motherboard ribbon-cable insertion (bottom center), and the 10-trajectory rolling success rate versus the number of training trajectories for both tasks (right). The shaded region marks the warm-up phase.}
\label{fig:real-world-learning-curves}
\end{figure*}

We next examine whether the gains from task-specific action-relevant routing extend to physical interaction. Figure~\ref{fig:real-world-learning-curves} shows the robot setup, representative rollouts, and learning curves, while Table~\ref{tab:real-world} reports normalized AUC and final-window success rate.

\noindent\textbf{Quantitative results.}
On USB connector insertion, eRLT achieves an AUC of $0.562$, exceeding RLT by $0.293$ ($108.9\%$ relative), and raises final-window success from $50\%$ to $90\%$. On ribbon-cable insertion, eRLT increases AUC from $0.484$ to $0.710$, an absolute gain of $0.226$ ($46.7\%$ relative), while both eRLT and RLT reach $100\%$ final-window success. DSRL$^\ast$ is less reliable on both tasks. These results show that eRLT improves the rate of online learning and, on the more difficult USB task, the final reliability of the policy.

\begin{table}[H]
\caption{\textbf{Real-world results.} Normalized AUC and final-window success rate (SR) under a fixed task-specific online interaction budget.}
\label{tab:real-world}
\begin{center}
\footnotesize
\setlength{\tabcolsep}{4pt}
\begin{tabular}{lcccc}
\toprule
Method & \multicolumn{2}{c}{USB Connector} & \multicolumn{2}{c}{Ribbon Cable} \\
\cmidrule(lr){2-3}\cmidrule(lr){4-5}
& AUC $\uparrow$ & SR (\%) $\uparrow$ & AUC $\uparrow$ & SR (\%) $\uparrow$ \\
\midrule
Frozen VLA & -- & 30.00 & -- & 70.00 \\
DSRL$^\ast$ & 0.266 & 40.00 & 0.272 & 50.00 \\
RLT & \underline{0.269} & 50.00 & \underline{0.484} & 100.00 \\
\textbf{eRLT (ours)} & $\mathbf{0.562}$ & 90.00 & $\mathbf{0.710}$ & 100.00 \\
\bottomrule
\end{tabular}
\end{center}
\end{table}
\vspace{-15pt}

\FloatBarrier

\noindent\textbf{Failure analysis.}
Inspection of unsuccessful rollouts helps explain the performance gap. RLT improves early but often retains a small offset between the connector and target interface. Although visually subtle, this residual error is sufficient to prevent insertion upon contact. DSRL$^\ast$ produces less stable corrections and leaves larger positional or angular errors. eRLT more consistently corrects these offsets before contact, while its remaining failures still involve fine alignment near the interface. Thus, task-specific action-relevant routing primarily improves the precision and consistency of the critical alignment phase.

\section{Conclusion and Limitations}

We presented eRLT, which constructs a task-specific RL token for lightweight online RL with a frozen VLA. eRLT uses learned routing tokens to aggregate task-specific action-relevant information across prefix tokens and a layer router to combine the resulting information across VLA depths. The routing module is initialized through expert action prediction and refined by the critic's Bellman objective. Across LIBERO and RoboTwin, eRLT outperforms the reconstruction-based RLT token on all seven tasks and the independent DSRL encoder on six of seven tasks in normalized learning-curve AUC, indicating improved sample efficiency. Ablations further show that token routing, layer routing, offline initialization, and online adaptation make complementary contributions. Results on two real-world insertion tasks demonstrate improved sample efficiency and, on USB connector insertion, a higher final-window success rate under a limited physical-interaction budget.

eRLT is limited by the information encoded in the frozen VLA, and its multi-layer extraction and periodic RL token recomputation add computation and memory relative to a fixed final-layer RL token. The real-world evaluation covers only two insertion tasks on one robot under a human-in-the-loop protocol; run-to-run variation and sensitivity to intervention frequency therefore remain insufficiently characterized. Future work should study router transfer across tasks and embodiments, additional backbones and robots, and longer-horizon tasks.

\bibliography{references}
\bibliographystyle{\paperbibliographystyle}

\clearpage
\appendix
\onecolumn

\section{Lightweight Online RL with a Frozen VLA}
\label{app:lightweight-online-rl}

\begin{figure}[h]
\centering
\includegraphics[width=\textwidth, trim={0.5mm 0.5mm 0.5mm 0.5mm}, clip]{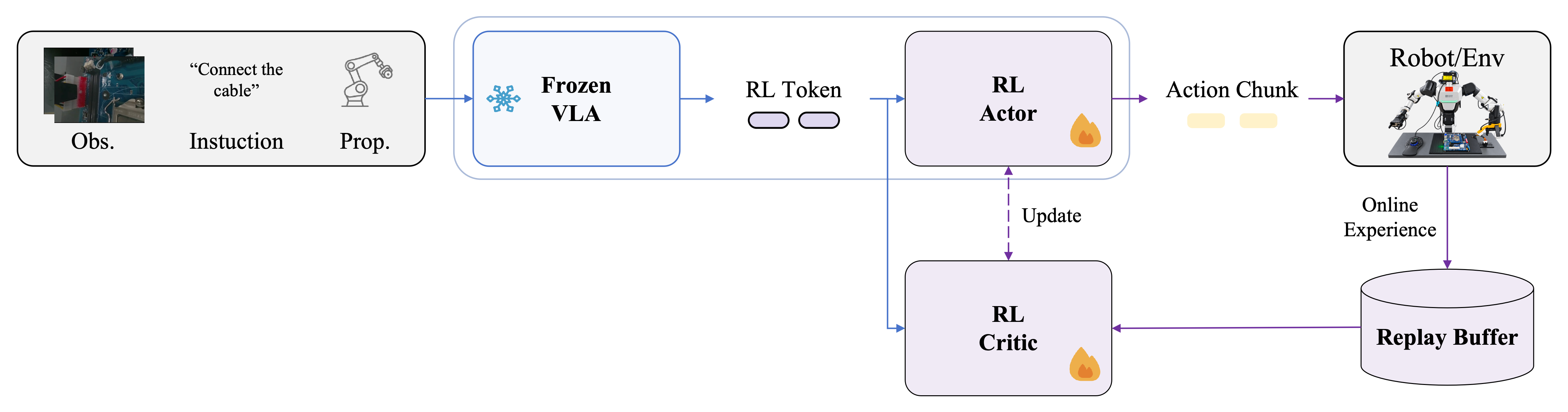}
\caption{\textbf{Lightweight online RL with a frozen VLA.} A frozen VLA and a lightweight actor jointly generate executable action chunks. Online interactions train the lightweight critic and actor, and the RL token supplies task-specific action-relevant information to both modules.}
\label{fig:lightweight-online-rl}
\end{figure}

Figure~\ref{fig:lightweight-online-rl} summarizes the interaction and learning loop used by lightweight online RL. After the combined policy executes an action chunk, the resulting transition is stored in a replay buffer and reused for off-policy actor-critic updates, while the pretrained VLA remains fixed.

\section{Algorithmic Details of eRLT}
\label{app:routing-optimization}

\subsection{Routing Architecture}

\paragraph{Routing path.}
The routing tokens are used only to construct the RL token and do not affect VLA action generation. We append them after the multimodal prefix and process the resulting sequence with the frozen VLM in a separate forward pass. Each routing token can attend to the original visual-language tokens, while the original tokens cannot attend to the appended routing tokens. We collect the routing-token hidden states at all selected layers from this same forward pass. For $K$ routing tokens and $m$ selected layers, mean pooling over the $K$ routing-token positions produces $m$ summaries in $\mathbb{R}^{D}$, and the layer router combines them into one RL token $z_t\in\mathbb{R}^{D}$. The routing module adds only $KD+m$ effective trainable parameters: $KD$ routing-token parameters and $m$ scalar layer logits.

\paragraph{Token layout and tensor shapes.}
For the $\pi_{0.5}$ backbone used in this work, the prefix tensor contains three image slots with 256 tokens per slot followed by 200 language-token positions, giving 968 positions in total, with hidden width $D=2048$. The attention mask determines which positions are valid according to the available camera views and language input. Appending $K$ routing tokens produces a sequence with $968+K$ positions. The complete tensor flow is shown in Table~\ref{tab:erlt-tensor-shapes}.

\begin{table}[h]
\centering
\small
\caption{Tensor shapes for the $\pi_{0.5}$ eRLT routing module. $B$ denotes batch size.}
\label{tab:erlt-tensor-shapes}
\begin{tabular*}{\textwidth}{@{\extracolsep{\fill}}lll@{}}
\toprule
Quantity & Shape & Construction \\
\midrule
Frozen multimodal prefix & $B\times968\times2048$ & $3\times256$ image slots $+200$ language positions \\
Routing input & $B\times(968+K)\times2048$ & Prefix followed by $K$ routing tokens \\
Per-layer routing-token states & $B\times K\times2048$ & Last $K$ positions at one selected index \\
Per-layer summaries & $B\times m\times2048$ & Mean over routing tokens, then stack $m$ indices \\
Layer weights & $m$ & Softmax of task-level logits with temperature $\tau$ \\
RL token $z_t$ & $B\times2048$ & Weighted sum over selected indices \\
\bottomrule
\end{tabular*}
\end{table}

The layer logits are input independent but task specific, whereas the routing-token states depend on the current images and instruction.

\subsection{Offline Action-Relevant Initialization}

During offline initialization, the action-prediction loss updates the routing tokens, layer logits, and temporary action probe. The frozen VLA is traversed by the gradient only to update the appended routing tokens; none of its parameters change. The action probe is discarded after initialization.

\paragraph{Algorithmic specification.} 
The implementation performs the following steps:
\begin{enumerate}
    \setlength{\itemsep}{1pt}
    \setlength{\parskip}{0pt}
    \item Sample demonstration frames and construct the following $H$-step expert action chunk, repeating the final action only when a chunk reaches an episode boundary. Apply the same action normalization used by the frozen VLA, producing an $Hd_a$-dimensional target for action dimension $d_a$.
    \item Run the frozen image and language embedding path, append the routing tokens, and collect their hidden states at the selected layers. Mean-pool the $K$ routing tokens at each layer and combine the resulting summaries with the learned softmax layer weights.
    \item Pass the RL token through the temporary action probe. Optimize the routing-token embeddings, layer logits, and probe jointly with the action-chunk MSE in Equation~\ref{eq:action-proxy-loss}; all VLA parameters remain frozen.
    \item Evaluate action MSE periodically, retain the routing parameters with the lowest validation MSE, and discard the probe before online RL.
\end{enumerate}

\subsection{Critic-Driven Online Adaptation}

During online learning, the critic loss updates the critic and routing parameters, while the actor uses detached RL tokens. The critic parameters remain unchanged during an actor update, although the derivative of its output with respect to the actor decision is retained for optimizing the actor. The Bellman target is computed using target copies of the actor, critic, and routing module. Because the RL token changes online, replay stores the raw current and next observations rather than precomputed RL tokens.

\paragraph{Algorithmic specification.}
Each learner iteration performs the following steps:
\begin{enumerate}
    \setlength{\itemsep}{1pt}
    \setlength{\parskip}{0pt}
    \item Sample transitions from replay and compute the target RL token for the next observation under no gradient. Use the target actor and critic to construct the controller-specific Bellman target.
    \item Update the critic using the RL tokens and update the actor using detached RL tokens. Keep the routing parameters fixed on ordinary learner iterations.
    \item Every $U$ learner iterations, recompute the current RL token with gradients enabled and jointly update the critic, routing-token embeddings, and layer logits from Equation~\ref{eq:critic-routing-loss}.
    \item Polyak-update the target actor, critic, and routing module toward their online counterparts.
\end{enumerate}

\section{Complete Experimental Setup}
\label{app:experimental-setup}

\subsection{Simulation Benchmarks}

\begin{figure}[h]
\centering
\includegraphics[width=\textwidth, trim={0.5mm 0.5mm 0.5mm 0.5mm}, clip]{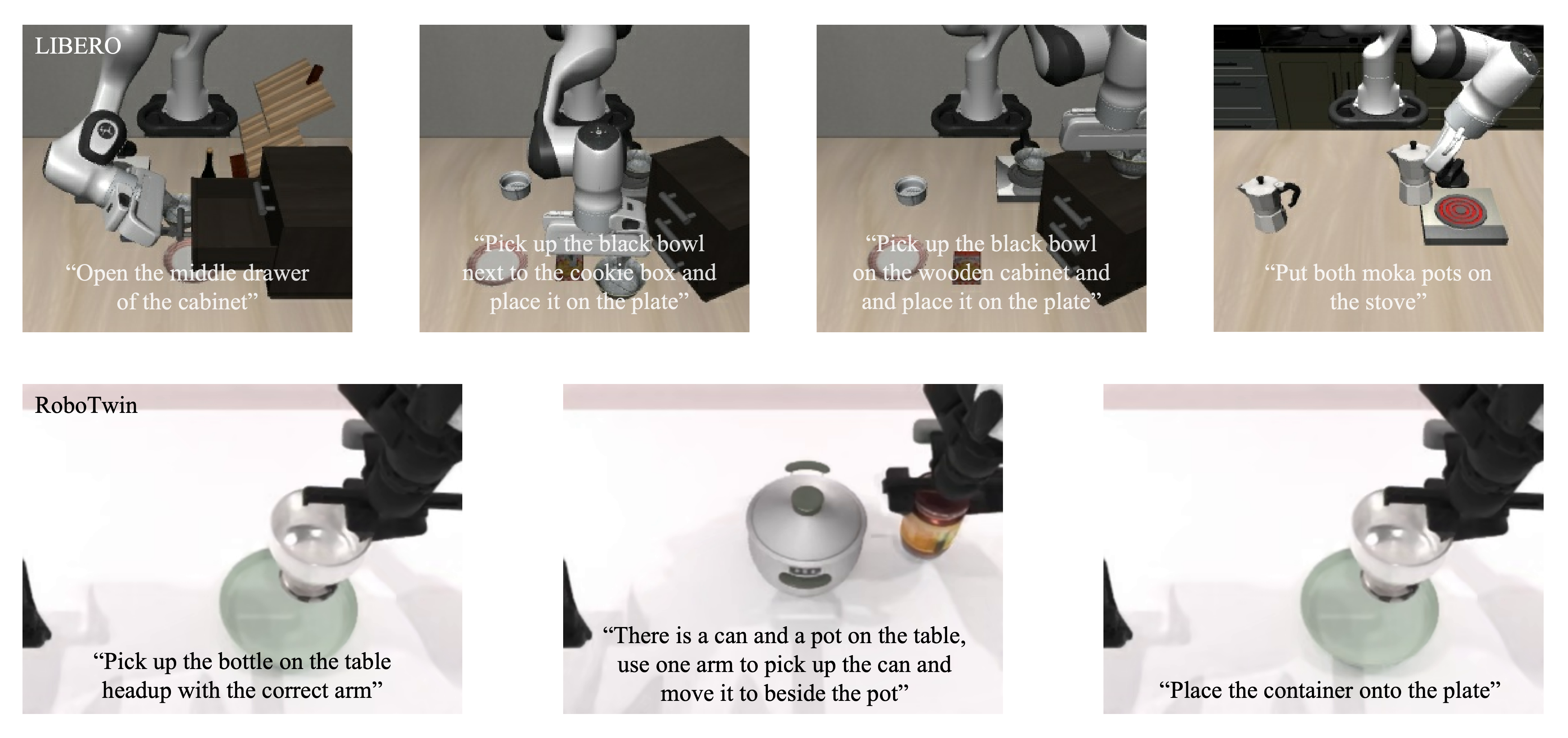}
\caption{\textbf{Simulation task overview.} The four LIBERO tasks (top) and three RoboTwin tasks (bottom) used in our simulation experiments.}
\label{fig:simulation-task-overview}
\end{figure}

For LIBERO, we focus on four individual tasks for which the frozen base VLA leaves substantial room for improvement: Drawer Opening, Bowl-to-Plate, Cabinet-Bowl-to-Plate, and Two-Moka-Pots. For RoboTwin, we evaluate on Adjust Bottle, Move Can--Pot, and Place Container Plate. Figure~\ref{fig:simulation-task-overview} visualizes all seven tasks.

Table~\ref{tab:simulation-task-protocol} gives the task identities used by the launch configurations. LIBERO assigns 50 benchmark initial states to each task. Training samples these states randomly with replacement, whereas evaluation uses 100 environments in deterministic order, traversing the 50-state pool twice. LIBERO episodes are capped at 240 low-level steps. RoboTwin instead follows its per-task environment configurations: 200 steps for Adjust Bottle and Move Can--Pot, and 150 steps for Place Container Plate. The reward is $-1$ on a non-success step and $0$ on the success step, obtained from a $-1$ step cost and a $+1$ success bonus. Success is the corresponding benchmark task predicate, and the reported success rate records whether it becomes true at least once before the task-specific horizon. Each LIBERO environment provides a $256\times256$ base-camera image, a $256\times256$ wrist image, and the 8-dimensional end-effector state. RoboTwin provides a head-camera view and two wrist-camera views at an environment resolution of $640\times480$; all three are resized to $224\times224$ for the VLA. Its 14-dimensional proprioceptive state contains six joint coordinates and one gripper coordinate for each arm.

\begin{table}[h]
\centering
\footnotesize
\setlength{\tabcolsep}{4pt}
\caption{Simulation tasks, horizons, and frozen-VLA success rates.}
\label{tab:simulation-task-protocol}
\begin{tabularx}{\textwidth}{@{}>{\raggedright\arraybackslash}p{0.27\textwidth}>{\raggedright\arraybackslash}X>{\raggedright\arraybackslash}p{0.10\textwidth}>{\centering\arraybackslash}p{0.08\textwidth}>{\centering\arraybackslash}p{0.08\textwidth}@{}}
\toprule
Task (suite) & Language instruction & Initial states & Horizon & Frozen SR \\
\midrule
\multicolumn{5}{@{}l}{\textit{LIBERO}} \\
Drawer Opening (Goal) & Open the middle drawer of the cabinet. & 0--49 & 240 & 0.74 \\
Bowl-to-Plate (Spatial) & Pick up the black bowl next to the cookie box and place it on the plate. & 300--349 & 240 & 0.78 \\
Cabinet-Bowl-to-Plate (Spatial) & Pick up the black bowl on the wooden cabinet and place it on the plate. & 450--499 & 240 & 0.12 \\
Two-Moka-Pots (LIBERO-10) & Put both moka pots on the stove. & 400--449 & 240 & 0.14 \\
\addlinespace[2pt]
\multicolumn{5}{@{}l}{\textit{RoboTwin}} \\
Adjust Bottle & Adjust bottle. & Seed list & 200 & 0.60 \\
Move Can--Pot & Move can to pot. & Seed list & 200 & 0.12 \\
Place Container Plate & Place container on plate. & Seed list & 150 & 0.72 \\
\bottomrule
\end{tabularx}
\end{table}

\subsection{Simulation Implementation and Training Protocol}
\label{app:simulation-implementation}

In simulation, the DSRL-style controller steers the frozen VLA in its latent action space. All variants within each benchmark share the same VLA checkpoint, observations, actor and critic architectures, replay data, reward definition, update schedule, and interaction budget. Unless otherwise stated, eRLT uses one learned routing token.

\paragraph{Frozen VLA and action parameterization.}
For LIBERO, we use the publicly released \href{https://huggingface.co/RLinf/RLinf-Pi0-LIBERO-Spatial-Object-Goal-SFT}{RLinf-Pi0-LIBERO-Spatial-Object-Goal-SFT} checkpoint for both rollout and training. The environment supplies two RGB views and an 8-dimensional state formed by end-effector position, end-effector orientation in axis--angle form, and gripper joint positions. The VLA predicts 10 delta actions with seven environment dimensions. The lightweight actor does not directly regress these actions: it samples a 32-dimensional Gaussian noise vector, repeats it across the 10-step flow horizon, and the frozen $\pi_{0.5}$ model converts it to the normalized action chunk using three integration steps. All ten resulting actions are sent to the environment before the next policy query. The RoboTwin experiments analogously use the publicly released \href{https://huggingface.co/motus-robotics/pi0.5_robotwin2}{pi0.5\_robotwin2} checkpoint. For both benchmarks, the demonstration data used for the offline initialization described below are the same data used to SFT the corresponding VLA checkpoint. The RoboTwin VLA predicts a 32-step sequence of 14-dimensional dual-arm joint-and-gripper actions, all 32 of which are executed before the next policy query; the lightweight actor again operates in a 32-dimensional latent action space. Episode termination follows the task-specific horizons in Table~\ref{tab:simulation-task-protocol}.

\paragraph{eRLT routing module and offline initialization.}
LIBERO and RoboTwin use the same routing configuration: $D=2048$, one routing token, hidden-state indices $\{0,3,6,9,12,15,18\}$, and a layer-router temperature of 1.0. Routing tokens are initialized from $\mathcal{N}(0,0.02^2)$ before offline training. The implementation allocates a fixed 64-entry logit vector and slices its first seven entries; the remaining entries receive no gradient. For $K=1$, the routing module therefore has 2,112 allocated parameters, of which 2,055 are active.

For the LIBERO initializer, each sampled frame is paired with the following 10-step expert action chunk, with the final action repeated only at an episode boundary. Applying the VLA action normalization and retaining the seven environment-action dimensions produces a 70-dimensional regression target. The temporary action probe has widths $2048$--$256$--$256$--$70$, with LayerNorm and ReLU in its two hidden layers. The initializer reuses the same post-training/SFT data used to obtain the frozen VLA checkpoint; the inspected subset contains 40 demonstration episodes. A seeded 5\% frame-level holdout produces 6,300 training frames and 332 validation frames. We train for 5,000 steps with batch size 2 using AdamW, learning rate $10^{-3}$, weight decay $10^{-4}$, and gradient-norm clipping at 1. Validation is run every 250 steps. The split, routing-token initialization, and minibatch sampler use seed 123. The routing-token-capacity ablation repeats this procedure independently for $K\in\{1,2,4,8\}$ while retaining a 2048-dimensional RL token through mean pooling over routing tokens.

For the RoboTwin initializer, each sampled frame is paired with the following 32-step, 14-dimensional expert action chunk, producing a 448-dimensional regression target. The corresponding probe has widths $2048$--$256$--$256$--$448$. The initialization dataset contains 50 clean demonstration trajectories for each of the 50 RoboTwin tasks, for 2,500 trajectories in total. Within each task, 45 trajectories are used for training and five for validation. Sampling frames with stride 8 produces 62,799 training frames and 7,008 validation frames. We make one pass over the training frames, corresponding to 31,400 steps with batch size 2, using the same optimizer, learning rate, weight decay, and gradient clipping as for LIBERO. Validation is run every 5,000 steps, the training and sampling seed is 0, and the routing parameters with the lowest validation MSE are retained.

\begin{table}[h]
\centering
\footnotesize
\caption{Offline action-relevant initialization used in simulation.}
\label{tab:simulation-offline-initialization}
\begin{tabularx}{\textwidth}{@{}lXX@{}}
\toprule
Configuration & LIBERO & RoboTwin \\
\midrule
Demonstration data & 40 episodes & 2,500 trajectories from 50 tasks \\
Validation split & 5\% of frames & 5 of 50 trajectories per task \\
Train/validation frames & 6,300 / 332 & 62,799 / 7,008 \\
Action target & $10\times7=70$ & $32\times14=448$ \\
Probe & $2048$--$256$--$256$--$70$ & $2048$--$256$--$256$--$448$ \\
Optimizer & AdamW & AdamW \\
Learning rate / weight decay & $10^{-3}$ / $10^{-4}$ & $10^{-3}$ / $10^{-4}$ \\
Batch size / training steps & 2 / 5,000 & 2 / 31,400 \\
Gradient clip & 1.0 & 1.0 \\
Checkpoint selection & Minimum validation MSE & Minimum validation MSE \\
\bottomrule
\end{tabularx}
\end{table}

\paragraph{Online RL architecture and objectives.}
For eRLT and RLT$^\ast$, the actor receives the 2048-dimensional RL token directly. It applies linear--LayerNorm--ReLU blocks of widths $(1024,512,256)$ and predicts a state-dependent mean and log standard deviation for the 32-dimensional latent action. Its log standard deviation is initialized to zero and clamped so that the standard deviation lies in $[\exp(-20),\exp(2)]$. The critic contains ten independent Q-heads. Each concatenates the RL token and latent action, applies the same $(1024,512,256)$ block widths, and outputs one scalar. Both Bellman targets and actor updates use the mean of the ten heads.

Let $H$ denote the executed action-chunk horizon, with $H=10$ for LIBERO and $H=32$ for RoboTwin, and let $d_t$ indicate whether any low-level step in the chunk terminates the episode. The target uses the first low-level reward in the chunk and discounts the bootstrap across the full chunk. Since the entropy coefficient is fixed to zero and the ten target Q-heads are mean aggregated, both benchmarks use
\begin{equation}
y_t=r_{t,0}+\gamma^{H}(1-d_t)\frac{1}{10}
\sum_{j=1}^{10}\overline{Q}_j\!\left(
\operatorname{stopgrad}(z_{t+1}),\xi_{t+1}\right),
\qquad \gamma=0.999,
\label{eq:simulation-dsrl-target}
\end{equation}
where $\xi_{t+1}$ is sampled from the current actor under no gradient and $\overline{Q}$ denotes the target Q-ensemble. Each online Q-head minimizes MSE to $y_t$, and the actor minimizes the negative mean Q-value. No entropy term is added to either objective.

\paragraph{Online training and routing updates.}
Training uses 16 parallel environments on one node. Each online epoch collects one rollout per environment, capped by the benchmark-specific horizon in Table~\ref{tab:simulation-task-protocol}, and stores the raw observations, tokenized instruction, latent action, rewards, and terminations in the trajectory replay buffer. Training begins at online epoch 9. Each subsequent epoch performs 200 replay updates with a global batch size of 256; the replay cache and sampling window contain 15,000 trajectories.

The actor and critic are updated on every replay step. For the first 199 updates, the current- and next-state RL tokens are computed once per batch and reused in detached form, leaving the routing parameters fixed. On the final update, the current-state RL token is recomputed with gradients enabled, so the critic loss also updates the routing-token embeddings and layer logits; the actor and target paths remain detached. This yields one routing update per online epoch. Standard updates use micro-batches of 64, while the routing update uses micro-batches of one with gradient accumulation.

After every replay update, the complete target model is Polyak-updated with coefficient $0.005$, using a persistent float32 shadow for bfloat16 parameters. The actor, critic, and router use Adam with constant learning rates $10^{-4}$, $3\times10^{-4}$, and $3\times10^{-4}$, respectively. Actor gradients are clipped at 3.5 and critic and router gradients at 10. Evaluation is performed every 20 epochs and at the final epoch. The actor uses no-shard FSDP with gradient checkpointing disabled, and the separate optimizers retain the original parameter objects.

\begin{table}[h]
\centering
\small
\setlength{\tabcolsep}{4pt}
\caption{eRLT routing hyperparameters across experimental settings.}
\label{tab:routing-hyperparameters}
\begin{tabular*}{\textwidth}{@{\extracolsep{\fill}}lcccccc@{}}
\toprule
Setting & Layers & $D$ & $K$ & $\tau$ & Router interval $U$ & Router target rate \\
\midrule
LIBERO & $\{0,3,6,9,12,15,18\}$ & 2048 & 1 & 1.0 & 200 & 0.005 \\
RoboTwin & $\{0,3,6,9,12,15,18\}$ & 2048 & 1 & 1.0 & 200 & 0.005 \\
Real world & $\{0,3,6,9,12,15,18\}$ & 2048 & 1 & 1.0 & / & / \\
\bottomrule
\end{tabular*}
\end{table}

\subsection{Real-World Tasks and Hardware}
\label{app:real-world-details}

\paragraph{Tasks and hardware.}
We evaluate USB connector insertion and motherboard ribbon-cable insertion using the left arm of an RB-Y1 robot and two wrist-mounted Intel RealSense D405 RGB cameras operating at 30\,Hz. Each episode begins with the connector already grasped. Both scenes use fixed tabletop targets: the USB receptacle is mounted upright on a power-strip fixture, while the motherboard is secured flat on the work surface. The deployment stack communicates with the robot through ROS~2 and runs the frozen VLA, routing module, actor, and critic on a workstation with a 24-core Intel Core Ultra 9 285K processor, 64\,GB memory, and one NVIDIA RTX PRO 5000 Blackwell GPU. The two wrist views are resized to $224\times224$ before entering the VLA; the head-camera slot is masked for both task-specific checkpoints. A rollout typically lasts approximately 20\,s and can extend to 30\,s when human assistance is required.

At reset, each of the seven nominal left-arm joint targets is independently perturbed by Gaussian noise with zero mean and standard deviation $0.01$\,rad. This joint-space perturbation induces variation in both the position and orientation of the initial end-effector pose; the torso and uncontrolled right arm are not perturbed.

\paragraph{Success and reward.}
Outcomes are judged manually. USB success requires the plug to be inserted into the USB receptacle, and ribbon-cable success requires the cable to be inserted into the motherboard connector; an attempt that does not achieve the corresponding insertion is a failure. Because the launch configuration imposes no automatic wall-clock limit, the operator declares a timeout when a rollout exceeds approximately $1.5\times$ the usual task-completion time. The operator closes each episode with one of three labels: success, failure, or truncation. The corresponding replay trajectory receives a sparse terminal reward of one only for success and zero otherwise; success and failure are terminal, while truncation is recorded separately.

\subsection{Real-World Implementation and Training Protocol}
\label{app:real-world-implementation}

In the real-world experiments, we adopt the RLT online RL pipeline and vary only the state representation. Our RLT implementation is built on the open-source \href{https://github.com/RLinf/RLinf}{RLinf} project. All adapted policies use the same task-specific VLA checkpoint, observations, actor and critic architectures, replay data, reward definition, update schedule, and interaction budget. We compare DSRL$^\ast$, RLT, and eRLT; here RLT denotes the condition using the original RL token in the RLT pipeline, while the Frozen VLA before online adaptation is reported as a reference.

\paragraph{Frozen VLA and action parameterization.}
The two deployed task-specific $\pi_{0.5}$ checkpoints are initialized from the official $\pi_{0.5}$ base weights. Their SFT datasets contain 99 complete-task USB demonstrations and 800 complete-task ribbon-cable demonstrations, respectively; we do not otherwise tune the VLA during online RL.

The control interface is shared by the two tasks. At each query, the frozen VLA predicts a 30-step chunk and the online policy executes the first 15 steps synchronously at 30\,Hz. Both the reference and learned policies use a seven-dimensional left-arm end-effector layout $(x,y,z,r_x,r_y,r_z,g)$, with Euler angles for orientation. The lightweight actor predicts normalized state-relative pose deltas: values in $[-1,1]$ correspond to at most $0.01$\,m of translation and $0.03$\,rad of rotation per step. Each predicted pose is decoded relative to the current proprioceptive state and converted to an absolute robot command. The gripper channel is not learned and is fixed to the closed command $g=0$; the deployment adapter additionally applies a $3$\,mm closing offset to reference gripper commands. Action prefetching and temporal smoothing are disabled, so the next observation is recorded only after the current 15-step chunk boundary.

\paragraph{State representations and offline initialization.}
For each task, RLT and eRLT are initialized separately using demonstrations from the corresponding SFT dataset. The USB initializer uses all 99 demonstrations, while the ribbon-cable initializer uses a 195-episode subset. For both methods, each task-specific dataset uses the same 10-episode holdout selected with split seed 42, leaving 89 USB episodes and 185 ribbon-cable episodes for training.

RLT trains a task-specific two-layer, eight-head reconstruction transformer from every demonstration frame, using 10,000 updates for USB and 20,000 updates for ribbon cable. Each initializer uses an effective batch of 48 (micro-batch 6), AdamW with a cosine schedule from $2\times10^{-5}$ to $2\times10^{-6}$ after 1,000 warm-up steps, weight decay $10^{-10}$, and gradient-norm clipping at 1.

For eRLT, a task-specific routing module is trained using the same hidden-state indices, width, routing-token count, and router temperature as in simulation. Each initializer subsamples frames with stride 8 and makes one pass over its training frames with batch size 2, AdamW, learning rate $10^{-3}$, weight decay $10^{-4}$, and gradient-norm clipping at 1. The temporary $2048$--$256$--$256$--$210$ probe predicts a normalized $30\times7$ expert action chunk; the routing parameters with the lowest validation MSE are retained and the probe is discarded.

\paragraph{Online RL architecture and objectives.}
The actor concatenates the first 15 reference actions, the 2048-dimensional RL token, and the 7-dimensional proprioceptive state. A three-layer MLP with widths $(256,256,256)$ outputs a $15\times7$ normalized action chunk. During collection it uses a fixed Gaussian standard deviation of $0.02$. The critic is a twin-Q ensemble; each Q-function uses three 256-wide hidden layers and receives the RL token, proprioception, and candidate action chunk.

The entropy coefficient is zero. For the first 100 learner updates, the actor uses only the behavior-cloning objective with weight 5. The Q-objective weight then increases linearly from 0 to 0.1 over the following 100 updates, while the behavior-cloning weight remains 5. During training, the reference chunk is independently dropped for 30\% of samples; rollout inference always receives it.

\paragraph{Online training and data collection.}
The replay cache and sampling window retain the most recent 200 trajectories. Learning begins once the replay buffer contains 100 action-chunk transitions, at which point the learner performs an initial block of 150 updates. Each newly collected episode then schedules 16 additional updates, with at most 25 executed in one training call. The critic is updated on every learner step and the actor on every second step. The global batch and micro-batch sizes are both 64. The discount factor is $0.993$, and the Polyak coefficient is $0.005$. The actor and critic use constant learning rates of $3\times10^{-4}$ with gradient clipping at 10.

Each run uses a fixed budget of 80 trajectories for USB or 90 trajectories for ribbon cable, including approximately 15 successful human-in-the-loop warm-up trajectories on average, and takes approximately 20--25 minutes. After warm-up, the robot collects autonomous rollouts. Five consecutive failures trigger five additional successful episodes provided by the operator through a 3Dconnexion SpaceMouse, after which autonomous collection resumes. The number of assisted trajectories therefore depends on the autonomous policy's performance.

\subsection{Motivating-Study Protocol}
\label{app:motivating-study-details}

We use RL Token (RLT) \citep{xu2026rltoken} as a controlled interface for studying layer and token choices. RLT compresses a prescribed set of VLA tokens with a reconstruction objective and keeps the resulting RL token fixed during online RL. All variants use the same actor-critic architecture and online RL procedure, with one factor varied at a time: the VLA layer with all tokens retained, or the token subset with the final layer retained. We report mean normalized evaluation-success AUC over three seeds. In Figure~\ref{fig:rlt-feature-selection}, gray marks the Final/All baselines, while outlines mark each task's best mean and parentheses report its gain over the corresponding baseline.

For the layer study, the source is varied among hidden-state indices 3, 9, 15, and the final hidden state while retaining all 712 selected prefix positions. For the token study, the final hidden state is fixed and we compare all 712 positions against three 328-token subsets. Each subset always retains all 200 language positions and 128 image positions. The 128 positions are either uniformly spaced or selected by the action-relevance scores computed separately from the Drawer Opening or Bowl-to-Plate demonstrations. These scores use absolute gradient-times-activation of diffusion action MSE, normalized within each layer and camera view and averaged across layers; the score is averaged over diffusion times $\{0.2,0.5,0.8\}$.

Each RLT variant uses the same two-layer reconstruction pretraining described above and remains frozen online. Online comparisons run for 100 epochs with three seeds, evaluate every 20 epochs on 100 ordered trials, and differ only in the source layer or token selection.

\subsection{Evaluation Protocol and Metrics}

AUC is computed from the raw evaluation points by trapezoidal integration. If $s_i$ is the success rate measured at online epoch $e_i$, the normalized value is
\begin{equation}
\operatorname{AUC}_{\mathrm{norm}}
=\frac{1}{e_N-e_1}\sum_{i=1}^{N-1}
\frac{s_i+s_{i+1}}{2}(e_{i+1}-e_i).
\label{eq:normalized-auc}
\end{equation}
No extrapolation is added before the first or after the last available evaluation point. Reported multi-seed results first compute Equation~\ref{eq:normalized-auc} per run and then average the resulting AUCs. Smoothing is used only for the dark visualization curves in Figure~\ref{fig:main-learning-curves}; all tables and AUCs use unsmoothed evaluation values. For the real-world experiments, each learning curve is the trailing success rate over autonomous episodes, plotted against the chronological number of collected training trajectories and evaluated over the fixed 80-trajectory USB or 90-trajectory ribbon-cable budget as detailed in Section~\ref{app:real-world-details}.

\section{Further Analysis of Real-World Online Adaptation}
\label{app:real-world-adaptation-analysis}

\subsection{Autonomous and Assisted Trajectories}
\label{app:real-world-trajectory-composition}

Table~\ref{tab:real-world-episode-accounting} summarizes the resulting composition of each fixed-budget collection. On USB insertion, eRLT collects 66 autonomous trajectories with 46 successes, compared with 57 trajectories and 20 successes for DSRL$^\ast$ and 44 trajectories and 18 successes for RLT. It consequently requires only 14 assisted trajectories, versus 23 for DSRL$^\ast$ and 36 for RLT. On ribbon-cable insertion, eRLT collects 70 autonomous trajectories with 63 successes and 20 assisted trajectories. RLT collects 65 autonomous trajectories with 44 successes and 25 assisted trajectories, whereas DSRL$^\ast$ collects 53 autonomous trajectories with 19 successes and 37 assisted trajectories. These differences arise because the common intervention rule requests additional assistance after five consecutive autonomous failures.

\begin{table}[h]
\centering
\footnotesize
\setlength{\tabcolsep}{3pt}
\caption{Trajectory composition for each real-world experimental condition. Successes are counted over autonomous trajectories only.}
\label{tab:real-world-episode-accounting}
\begin{tabular*}{\textwidth}{@{\extracolsep{\fill}}llcccc@{}}
\toprule
Task & Method & Total & Autonomous & Assisted & Autonomous successes \\
\midrule
USB connector & DSRL$^\ast$ & 80 & 57 & 23 & 20 \\
USB connector & RLT & 80 & 44 & 36 & 18 \\
USB connector & eRLT & 80 & 66 & 14 & 46 \\
Ribbon cable & DSRL$^\ast$ & 90 & 53 & 37 & 19 \\
Ribbon cable & RLT & 90 & 65 & 25 & 44 \\
Ribbon cable & eRLT & 90 & 70 & 20 & 63 \\
\bottomrule
\end{tabular*}
\end{table}

\subsection{Online Optimization Dynamics}
\label{app:real-world-optimization-dynamics}

\begin{figure}[p]
\centering
\includegraphics[width=\textwidth]{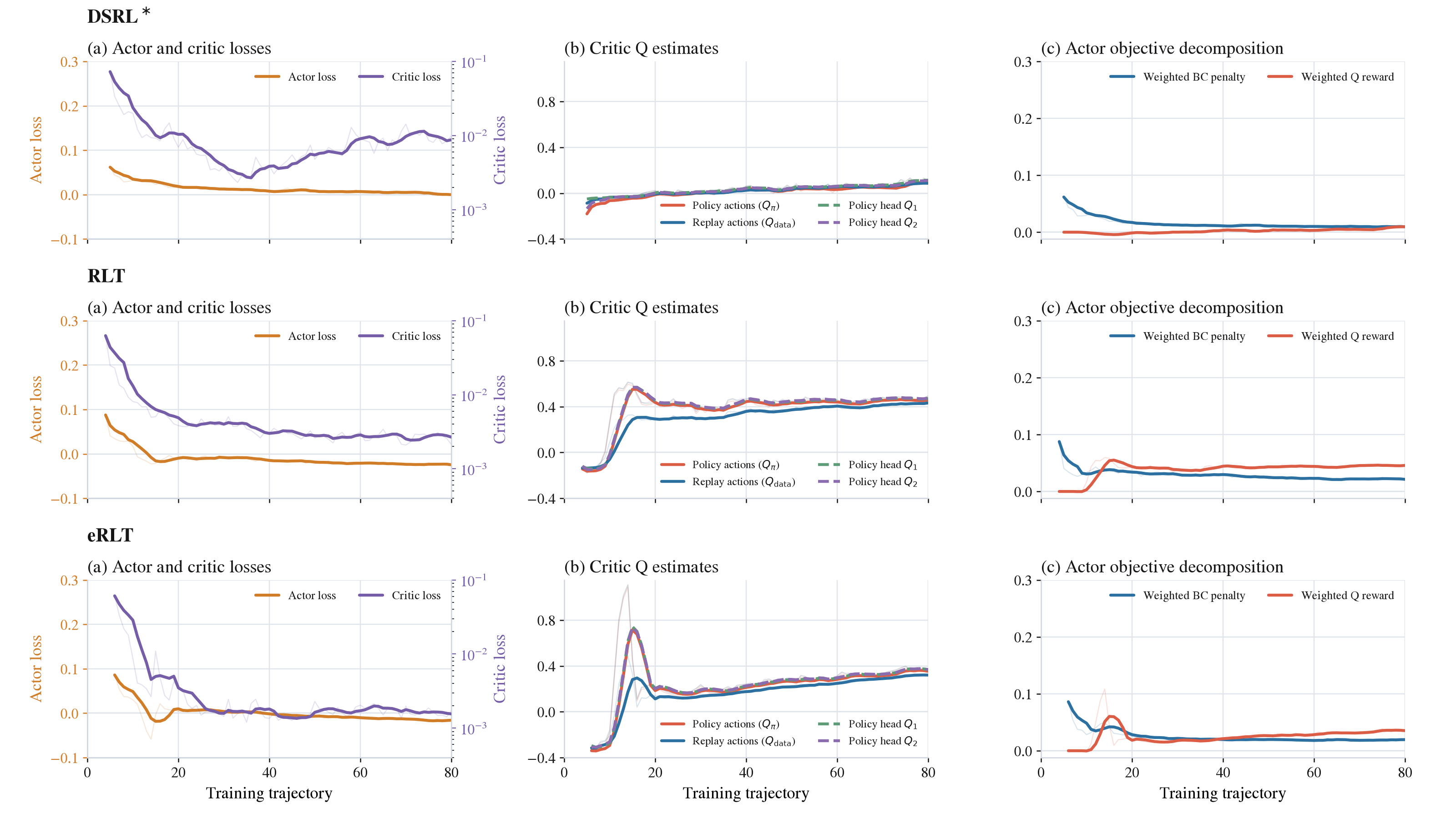}
\caption{Online optimization dynamics for USB connector insertion. Rows correspond to DSRL$^\ast$, RLT, and eRLT. Within each row, the panels show (a) actor and critic losses, (b) critic Q estimates for policy and replay actions, and (c) the weighted behavior-cloning and Q terms in the actor objective. Faint curves denote raw logged values and solid curves denote trailing-five-trajectory means.}
\label{fig:usb-training-diagnostics}
\end{figure}

\begin{figure}[p]
\centering
\includegraphics[width=\textwidth]{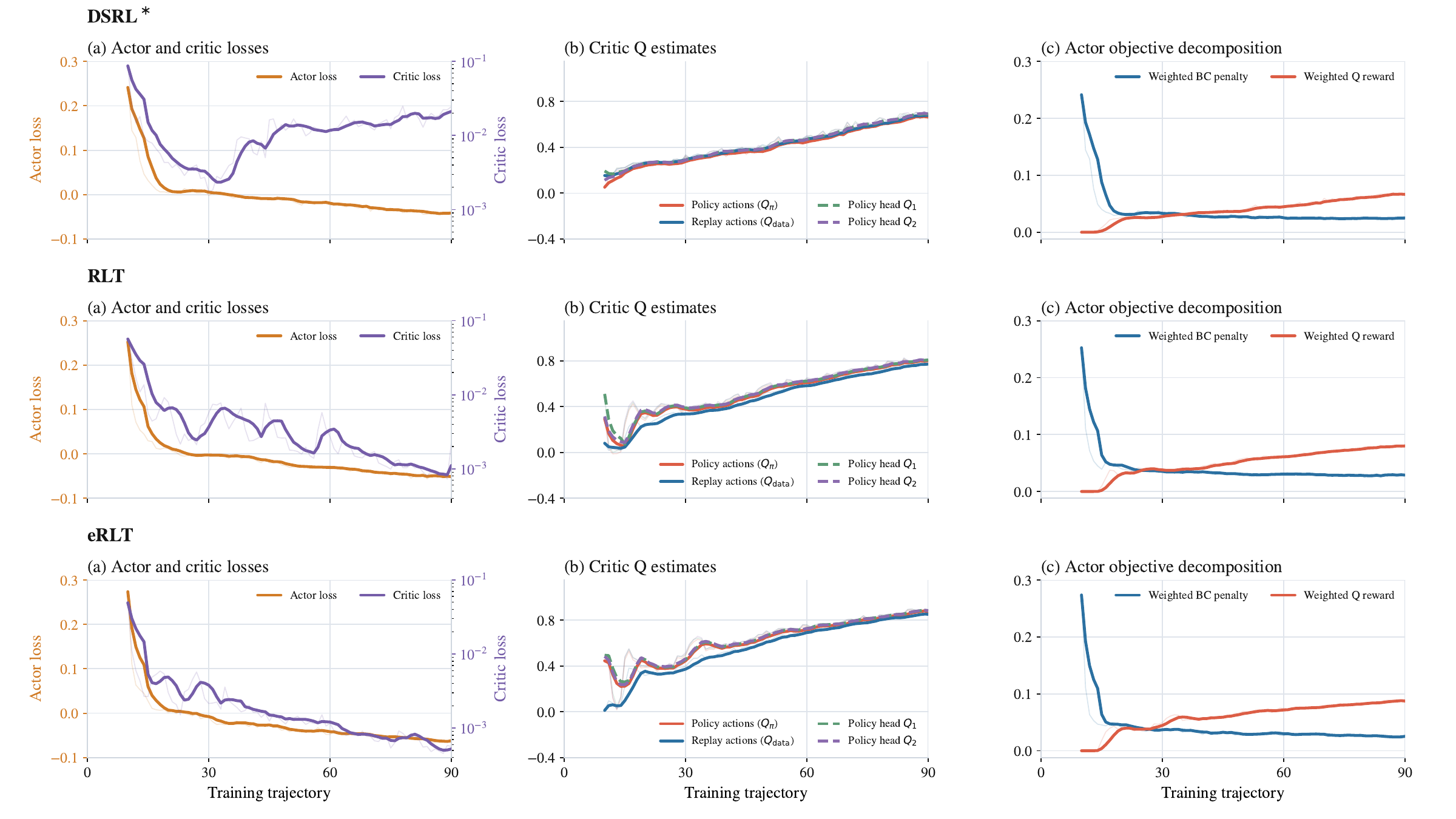}
\caption{Online optimization dynamics for ribbon-cable insertion. Rows correspond to DSRL$^\ast$, RLT, and eRLT, using the same metrics and smoothing convention as Figure~\ref{fig:usb-training-diagnostics}.}
\label{fig:cable-training-diagnostics}
\end{figure}

Figures~\ref{fig:usb-training-diagnostics} and~\ref{fig:cable-training-diagnostics} report the optimization trajectories for DSRL$^\ast$, RLT, and eRLT on the two real-world tasks. For each run, we show the actor and critic losses, the critic estimates for policy and replay actions, and the weighted behavior-cloning and Q terms in the actor objective. The horizontal axis denotes the chronological number of collected training trajectories. Faint curves show the raw logged values, while solid curves show trailing-five-trajectory means.

Across the six runs, the two critic heads remain closely aligned, and the Q estimates for policy and replay actions follow the same overall trend. The actor-objective curves also reflect the staged optimization described in Section~\ref{app:real-world-implementation}: the weighted behavior-cloning term decreases early in training, while the weighted Q term becomes active later and contributes increasingly to the actor update.

\end{document}